\documentclass[pdflatex,sn-vancouver-ay]{sn-jnl}

\usepackage{tabularx}%
\usepackage{array}
\usepackage{multirow}      
\usepackage{caption}       
\usepackage{graphicx}%
\usepackage{amsmath,amssymb,amsfonts}%
\usepackage{amsthm}%
\usepackage{lmodern}
\usepackage{mathrsfs}%
\usepackage[title]{appendix}%
\usepackage{xcolor}%
\usepackage{color,soul}
\usepackage{textcomp}%
\usepackage{manyfoot}%
\usepackage{booktabs}%
\usepackage{algorithm}%
\usepackage{lineno}
\usepackage{algorithmicx}%
\usepackage{algpseudocode}%
\usepackage{listings}%

\theoremstyle{thmstyleone}%
\theoremstyle{thmstyletwo}%

\theoremstyle{thmstylethree}%

\DeclareUnicodeCharacter{202F}{\,}

\begin{document}

\title[Article Title]{Improving watermelon~\textit{(Citrullus lanatus)} disease classification with generative artificial intelligence (GenAI)-based synthetic and real-field images via a custom EfficientNetV2-L model}


\author[1]{\fnm{Nitin} \sur{Rai}}

\author[1]{\fnm{Nathan S.} \sur{Boyd}}

\author[2]{\fnm{Gary E.} \sur{Vallad}}

\author*[3]{\fnm{Arnold W.} \sur{Schumann}}\email{schumaw@ufl.edu}

\affil[1]{\orgdiv{Gulf Coast Research and Education Center (GCREC)}, \orgname{University of Florida}, \city{Wimauma}, \state{FL}, \country{USA}}

\affil[2]{\orgdiv{Department of Plant Pathology, Gulf Coast Research and Education Center (GCREC)}, \orgname{University of Florida}, \city{Wimauma}, \state{FL}, \country{USA}}

\affil[3]{\orgdiv{Citrus Research and Education Center}, \orgname{University of Florida}, \city{Lake Alfred}, \state{FL}, \country{USA}}



\abstract{The current advancements in generative artificial intelligence (GenAI) models have paved the way for new possibilities for generating high-resolution synthetic images, thereby offering a promising alternative to traditional image acquisition for training computer vision models in agriculture. In the context of crop disease diagnosis, GenAI models are being used to create synthetic images of various diseases, potentially facilitating model creation and reducing the dependency on resource-intensive in-field data collection. However, limited research has been conducted on evaluating the effectiveness of integrating real with synthetic images to improve disease classification performance. Therefore, this study aims to investigate whether combining a limited number of real images with synthetic images can enhance the prediction accuracy of an EfficientNetV2-L model for classifying watermelon (Citrullus lanatus) diseases. The training dataset was divided into five treatments: H0 (only real images), H1 (only synthetic images), H2 (1:1 real-to-synthetic), H3 (1:10 real-to-synthetic), and H4 (H3 + random images to improve variability and model generalization). All treatments were trained using a custom EfficientNetV2-L architecture with enhanced fine-tuning and transfer learning techniques. Models trained on H2, H3, and H4 treatments demonstrated high precision, recall, and F1-score metrics. Additionally, the weighted F1-score increased from 0.65 (on H0) to 1.00 (on H3-H4) signifying that the addition of a small number of real images with a considerable volume of synthetic images improved model performance and generalizability. Overall, this validates the findings that synthetic images alone cannot adequately substitute for real images; instead, both must be used in a hybrid manner to maximize model performance for crop disease classification.}




\maketitle

\section{Introduction}\label{sec1}

Watermelon \textit{(Citrullus lanatus)} is considered one of the most important crops within the cucurbit family in terms of its nutritional content, particularly its high-water content and vitamins. According to the US Department of Agriculture (USDA) Economic Research Service, Florida accounted for 1.02 billion pounds of watermelons, surpassing Texas and California in production in 2021~\citep{weber_kramer_2022}. Such large-scale production requires consistent crop protection practices to minimize yield losses due to biotic stressors. It has been reported that growers typically apply fungicides every 7-14 days during watermelon production, averaging about four sprays per season~\citep{mossler2005watermelon}. Consistent fungicide applications facilitates season-long management of diseases, such as, such as gummy stem blight, powdery mildew, and anthracnose among others~\citep{keinath_miller_2022}. Smart spray technology is a tool that should facilitate reduced fungicide inputs with no loss of efficacy thereby providing early detection and farmer cost savings~\citep{ampatzidis_ufifas_2022,durham_usda_2016}. 

In Florida, watermelon is affected by multiple fungal diseases, such as anthracnose  \textit{(Colletotrichum obiculare)}, downy mildew \textit{(Pseudoperonospora cubensis)}, gummy stem blight \textit{(Didymella bryoniae)}, cercospora \textit{(Cercospora citrullina)}, and Alternaria leaf spots \textit{(Alternaria cucumerina)}~\citep{roberts2023anthracnose}. In addition, several types of viruses also affect the overall growth of this crop~\citep{ravikumara2025watermelon}. These pathogens affect different parts of the crop wherein the affected areas show a range of visible symptoms on vegetative tissues. For instance, anthracnose spreads to leaves, stems, and even the fruits and may cause leaking or decaying of fruit several days after harvest. On the contrary, downy mildew affects only the leaves and can cause leaf defoliation at later stages of the crop. In addition to these pathogens, viruses, such as watermelon mosaic virus can be spread by aphids and leaves show multiple symptoms, such as pale green color, dwarfing, or even show mosaic patterns of light and dark patches. While growers tend to apply fungicides to mitigate the effect of these diseases on watermelons, symptoms of two or more diseases can look similar especially early in the season. For instance, leaf damage caused by leaf miners \textit{(Liriomyza sativae)}, a common pest in watermelons, can be confused with anthracnose or gummy stem blight. 

Smart spray technology utilizes computer vision approach powered by deep learning (DL) models trained on extensive plant disease image datasets to identify targets of interest. These computer vision systems are integrated with pesticide sprayers enabling them to make on-the-go decisions and target spray as needed~\citep{li2022design,oberti2016selective}. A reliable computer vision system requires large-scale datasets and artificial intelligence (AI)-based classification algorithms trained on the datasets to enable classification of diseases in dynamic and uncontrolled environments. With regards to generating large-scale datasets for model training, the current research efforts are targeted towards using generative artificial intelligence (Gen AI) to create synthetic images of crop or weed datasets~\citep{lu2022generative,pallottino2025applications}. For instance, stable diffusion-based text-to-image approach was used to generate pest images in natural scenarios~\citep{wang2024stable}. Most researchers have relied on leveraging Generative Adversarial networks (GAN) for image-to-image translation in synthetic image generation. Over 15 different varieties of GAN networks have been used to generate synthetic images as part of image augmentation to train AI models for crop health assessment (Lu et al., 2022). Deep convolutional GAN (DCGAN) and Cycle-Consistent GAN (CycleGAN), are the most widely adopted approaches for synthetic image generation~\citep{douarre2019novel,gomaa2021early,zeng2020gans,sun2020data}.

In addition to the application of Gen AI for large-scale synthetic image generation, significant efforts have been put into developing AI models for disease classification. Various approaches ranging from straightforward classification to instance segmentation have been utilized to develop smart computer vision systems~\citep{li2022design,yang2025advanced}. These computer vision models can be subsequently be integrated with ground-based mapping or spray systems. Target spraying on diseases poses several challenges in terms of decision-making for real-time spraying strategy. First, some diseases exhibit similar symptomology, requiring further lab testing for verification. This process takes time and requires the expertise of a plant pathologist. Second, disease symptoms are often randomly scattered over the leaves which makes it computationally intensive task for the model to locate and draw bounding boxes. The end-goal in this case is just to identify the disease on a global image and execute the spraying operation without considering the lesion-level localization. In numerous studies, DL-based convolutional neural network (CNN) classification approaches have outperformed more complicated detection or instance segmentation techniques. For instance, a CNN architecture achieved 84.9\% accuracy in comparison to bounding-box-based You Only Look Once (YOLO) model in classifying pests and plant diseases~\citep{shoaib2025leveraging}. Another study used a CNN model to detect early-stage plant disease with high accuracy~\citep{venkateswara2025deep}. In consideration of the advancements made in the areas of large-scale synthetic crop image generation and CNN-based classification approaches to enhance real-time target spraying, not much effort has been allocated towards the efficacy of adding real-field images with synthetic images in enhancing crop disease classification. This research would answer an important question: Does adding real-field images in fusion with synthetic data lead to better disease classification? Therefore, the objectives of this research study are as follows:

\begin{enumerate}
    \item To evaluate the effectiveness of using only synthetic images as an alternative to manually collected real-field images for training deep learning models. 
    \item To determine if model accuracy improves with the inclusion of real images with synthetically generated images.
    \item To investigate the effect of adding irrelevant (unknown) classes on model prediction accuracy and classification performance.  
\end{enumerate}

Overall, Section~\ref{Sec2} discusses the methodology followed to create the data analysis pipeline starting from GenAI application for synthetic image generation to EfficientNetV2-L architecture customization for training and testing. Subsequently, Section~\ref{sec3} reports relevant results after deploying the model on a test dataset as per the objectives of this study.~Section~\ref{Sec4} provides critical discussion on the aspects of computer vision application for disease classification. In addition to this, Section~\ref{Sec5} highlights the limitations of the study along with future research directions. Finally, Section~\ref{Sec6} concludes the research work.    



\section{Materials and methods} \label{Sec2}

\begin{figure}[h!]
    \includegraphics[scale=0.39]{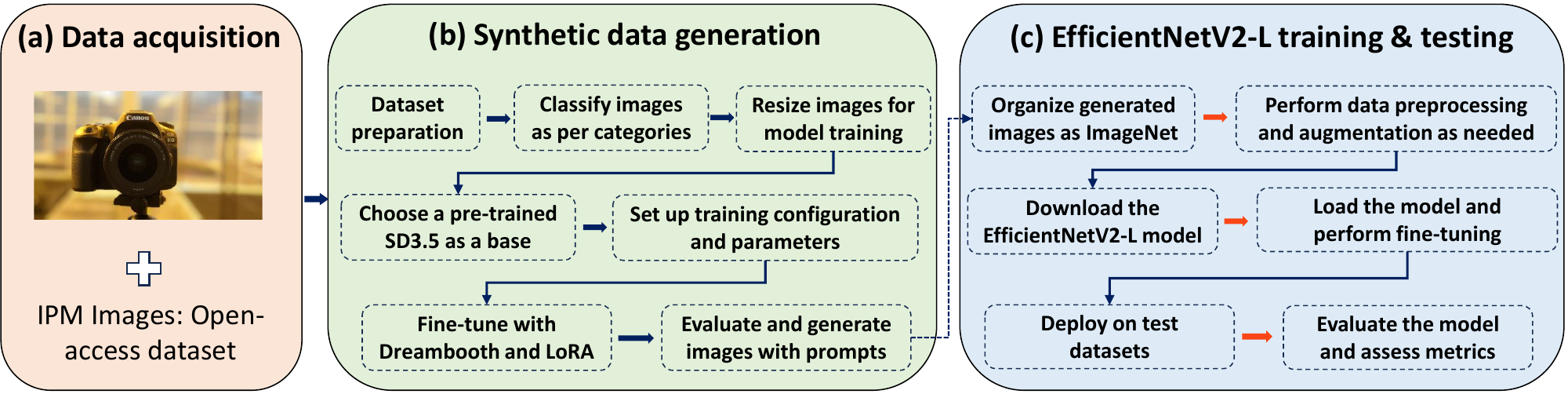}
    \vspace{-0.2cm}
    \caption{Overall illustration of systematic approach taken to perform watermelon disease classification right from image data generation to model training.}
    \label{Fig1}
\end{figure}

Figure~\ref{Fig1} displays the overview of the steps involved in performing this research study. In addition, model metrics are also explained subsequently in each section. 

\vspace{-1cm}

\subsection{Image data acquisition with expert advice} \label{sec2.1}

\begin{figure}[h]
    \centering
    \includegraphics[scale=0.39]{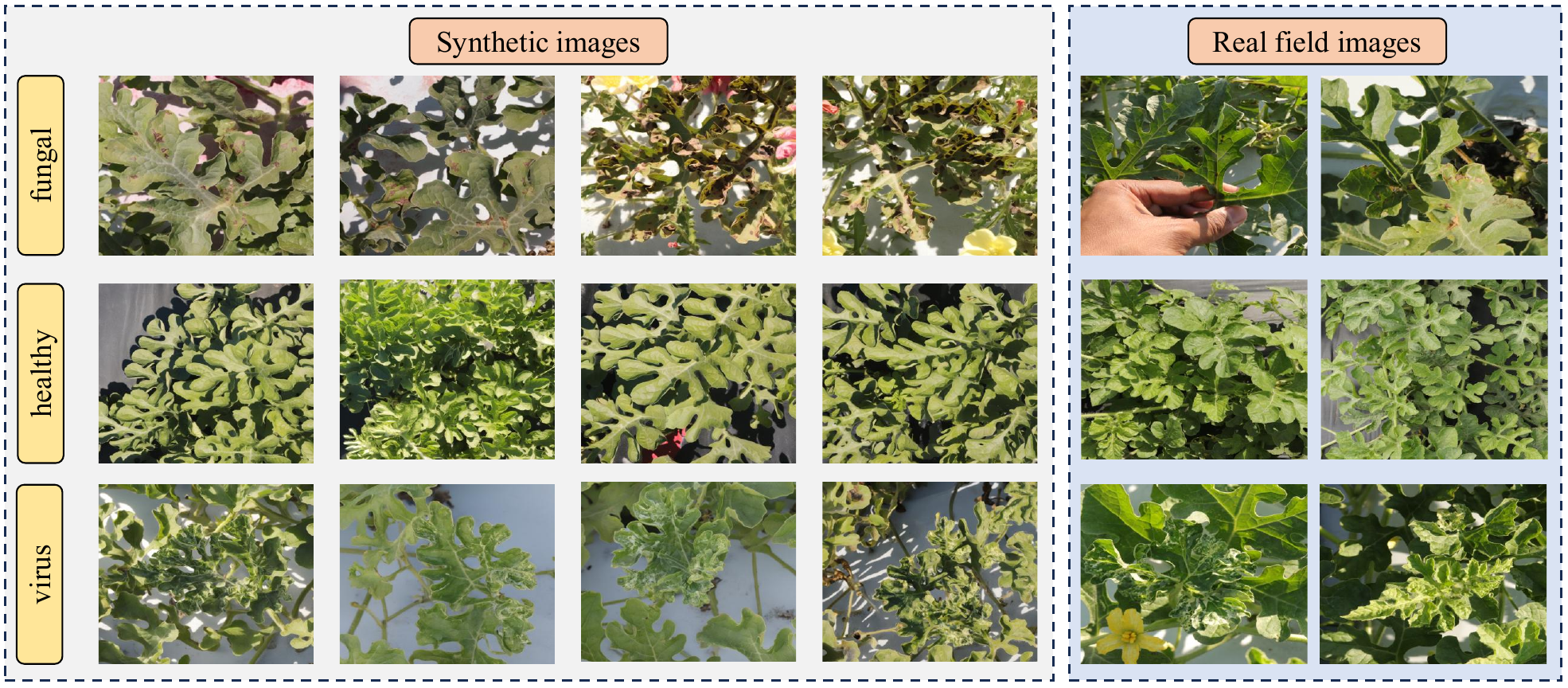}
    \vspace{-0.2cm}
    \caption{Sample images consisting of synthetic and real images used for model training and testing. Synthetic images consisting of anthracnose and downy mildew are circled with yellow and orange colors, respectively.}
    \label{fig2}
\end{figure}

The disease image dataset used for training and testing the model were labeled as three classes: fungal, healthy, and virus. The overall training set consisted of five treatments out of which the fifth treatment included a fourth class called unknown, taken from the ImageNet dataset~\citep{deng2009imagenet}. This unknown class was added so that the model would focus on features specific to plant diseases and disregard irrelevant patterns~\citep{robinson2021deep}. Within the fungal category, two diseases, anthracnose and downy mildew were used. For the virus class, images of watermelon mosaic virus were added to the training set. The dataset for fungal category, i.e., anthracnose and downy mildew were captured using Sony’s hand-held DSLR in the summer of 2025. The data acquisition was conducted from May 5\textsuperscript{th} to June 15\textsuperscript{th} at a resolution of 2048 $\times$ 1536. To prepare the test set for the analysis, multiple field trips were made to the experimental plots to capture images of all the classes. These images were further validated for specific disease under the guidance of an expert plant pathologist. Figure~\ref{fig2} shows the distinction between synthetically generated disease images and images captured in real-field conditions.

\vspace{-1cm}

\subsection{Overall image data construction for model training}
\subsubsection{Multi-modal generative AI for synthetic data generation}

Generative artificial intelligence (GenAI) is a type of artificial intelligence (AI) that is not limited to learning patterns in a single format, i.e., images, audios, or videos. On the contrary, the GenAI approach is capable of extracting patterns from a dataset developed from multiple sources to generate ``never seen before'' samples based on learned features. For instance, in smart agriculture, GenAI could be leveraged to generate synthetic images from a very limited sample of disease images. It is capable of learning or mimicking the diseased features from a set of sample images, and then generate synthetic versions with varying background and environmental conditions. It is capable of imparting diversity to the generated dataset, thereby rapidly producing thousands of synthetic images. This is specifically helpful in the agricultural domain since the training samples on the internet are less diverse and often not open-access leading to clustered and less robust classification model generation~\citep{han2025plant,lu2022generative,davies2009cluster}.

\begin{figure}[h]
    \centering
    \includegraphics[scale=0.39]{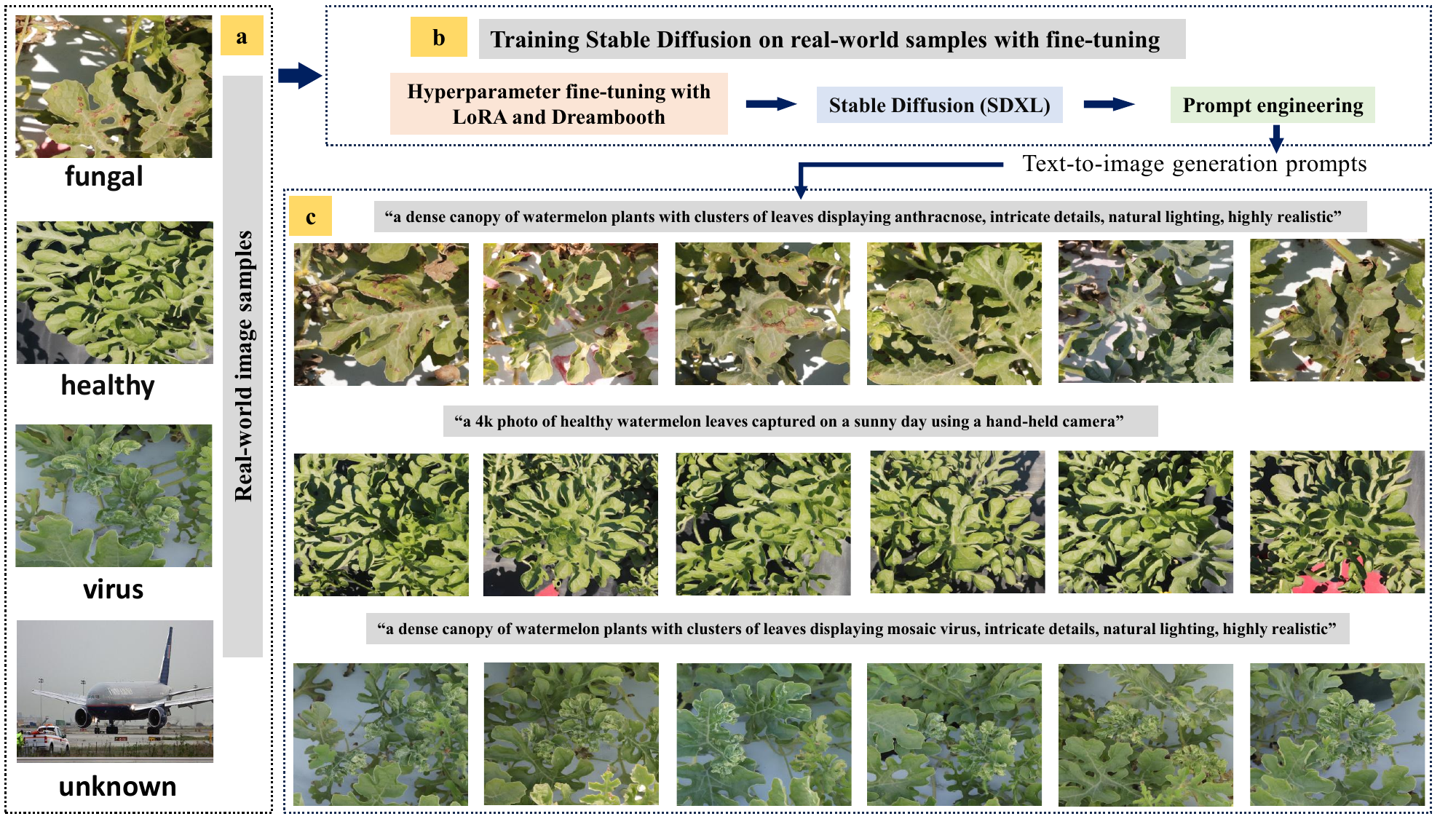}
    \vspace{-0.2cm}
    \caption{An illustration of steps taken to train Stable Diffusion (SD) generative model on real-world samples. It consists of, (a) gathering real-world samples, (b) training SD 3.5M using Low-Rank Adaption (LoRA) and Dreambooth fine-tuning platforms, and (c) applying prompt engineering to generate desired number of synthetic samples.}
    \label{Figure3}
\end{figure}

In this research, Stable Diffusion model (SD 3.5M) was trained with a few real images of multiple diseases caused by fungal and viral pathogens found in watermelons. In addition, two approaches, Learn on Reconstruction and Attention (LoRA)~\citep{hu2022lora} and DreamBooth~\citep{ruiz2023dreambooth}, were used to fine-tune the SD3.5M model to generate realistic looking images with varying background and lighting conditions. A minimum of 30 real images with best representations from a specific class were used to generate multiple synthetic images using a relevant prompt engineering approach. For the fungal class, 32 and 35 real images of anthracnose and downy mildew were used, respectively. Similarly, 34 images of healthy and 30 images of mosaic virus was used to generate synthetic images. During training the SD 3.5M model, diseases from individual classes were not mixed. For instance, only disease symptoms showing anthracnose were used to train and generate synthetic images of watermelon leaves with anthracnose. This strategy was adopted because it has been found that mixing multiple disease symptoms during SD3.5M model training could lead to the generation of unnatural synthetic images. After the model was trained, specific prompts were given to generate the synthetic images (Fig.~\ref{Figure3}c). Figure 3 displays samples of real images used, some fine-tuning approaches, and the prompts that were used to generate a diverse set of synthetic datasets. For more information on generating synthetic samples using the SD multi-modal model, refer to \cite{rai2025phytosynth}.

\vspace{-0.9cm}

\subsubsection{Disease dataset categorization and training platform}

The dataset was distributed into five treatments: (a) only real images (H0), (b) only synthetic images (H1), (c) real + synthetic1 (1:1) (H2), (d) real + synthetic2 (1:10) (H3), and H3 + random class (H4).  A random class (containing images from ImageNet) was added in the final treatment to prevent spurious results when anomalous images, such as plastic mulch, weeds or soil are fed to the model for classification. In such case, the model will classify it as “unknown” rather than labeling it as fungal, healthy or virus. The number of images in each class was balanced for all treatments to prevent bias and overfitting. Each treatment started with a fixed set of 750 real images, which was further split into training (70\%), validation (15\%), testing (15\%) sets. From this partition, 112 test images were set aside and kept constant across all treatments, while the remaining images were further split into training and validation sets (80\%:20\%). Table~\ref{Table1} displays the exact number of images used in each treatment for model training and testing. The overarching goal was to ensure fair comparison by controlling the overall dataset size allowing the effect of real image inclusion on model performance and generalization to be isolated without the influence of larger dataset size.~Furthermore, within the fungal class, two distinct fungal diseases, anthracnose, and downy mildew were merged into a single class for model training using EfficientNetV2-L. This grouping was done to simplify the classification task and reflect real-world disease diagnosis scenarios, where precise identification of a specific pathogen may not always be feasible. This approach enabled the model to distinguish fungal vs. non-fungal symptoms, which is often sufficient for triggering timely fungicide applications in practical settings.~The overall data analysis was carried on a high-performance computing system,~\href{https://ai.ufl.edu/research/hipergator/}{HiPerGator}, that was equipped with a single Nvidia B200 Quadro GPU with 180 GB of VRAM, 64 GB of RAM, and 10 CPU cores. Furthermore, Jupyter Notebook with Python (v3.11.11),~\href{https://docs.pytorch.org/vision/stable/index.html}{torchvision (v0.20.0)},~\href{https://pypi.org/project/torch/}{torch (v2.5.1)}, and~\href{https://developer.nvidia.com/cuda-12-4-0-download-archive}{CUDA 12.4} support was used to execute model training and testing.

\setlength{\tabcolsep}{5pt}  
\renewcommand{\arraystretch}{0.97} 
\begin{sidewaystable}
\caption{Overview of watermelon disease images used per class to train and test EfficientNetV2-L model.} \label{Table1}
\begin{tabularx}{\textwidth}{@{}p{1.3cm} p{3.0cm} p{1.8cm} p{1.8cm} p{2.5cm} p{1.8cm} X@{}}
\toprule%
Classes & Total images & Training & Validation & Testing & Split ratio & Purpose \\
\midrule
\multicolumn{7}{c}{\textit{H0 (only real images)}} \\
\hline
fungal & & & & & & \\
healthy & 750 (real) & 526 & 112 & 112 \emph{(Kept constant across all the treatments)} & 70\%:15\%:15\% & Baseline with 100\% real images \\
virus &&&&&& \\ 
\midrule
\multicolumn{7}{c}{\textit{H1 (only synthetic images)}} \\
\midrule
fungal &&&&&& \\
healthy & 638 (synthetic) & 510 & 128 & --- & 80\%:20\% & Baseline with 100\% synthetic images \\
virus &&&&&& \\
\midrule
\multicolumn{7}{c}{\textit{H2 (1real: 1synthetic)}} \\
\midrule
fungal &&&&&& \\
healthy & 638 (real) + 638 (synthetic) & 1,021 & 255 & --- & 80\%:20\% & Moderate number of synthetic images \\
virus &&&&&& \\
\midrule
\multicolumn{7}{c}{\textit{H2 (1real: 10synthetic)}} \\
\midrule
fungal &&&&&& \\
healthy & 638 (real) + 6,380 (synthetic) & 5,614 & 1,404 & --- & 80\%:20\% & Large number of synthetic images \\
virus &&&&&& \\
\botrule
\end{tabularx}
\end{sidewaystable}

\subsection{Proposed EfficientNetV2-L classification architecture with classifier head fine-tuning and transfer learning}

The EfficientNetV2-L classification architecture is a high-performance convolutional neural network (CNN) optimized for efficient training and testing tasks. In comparison to other state-of-the-art (SoTA) models, such as You Only Look Once (YOLO), EfficientNetV2-L model has four-fold advantages in real-time disease classification: (a) speed of the model, (b) robustness in understanding global features rather than individual disease symptoms, (c) no hand-annotated data labeling, and (d) no location (coordinate or bounding boxes) specific for faster spraying and actuation. However, EfficientNetV2-L pretrained models from the internet are trained on the ImageNet dataset which is only good to classify general domains with high precision and accuracy. Therefore, there is a need to train the model with the disease dataset in an agricultural domain, with additional fine-tuning approaches. As an overarching approach, this study uses a transfer learning technique that is accomplished in two steps: (a) training a new classifier head with additional layers, and (b) fine-tuning on the remaining layers (deeper layers) by selectively freezing base layers. Additionally, several hyperparameters to enhance the training strategy have been used, such as optimizer and callback techniques. 

\begin{figure}[h]
    \centering
    \includegraphics[scale=0.39]{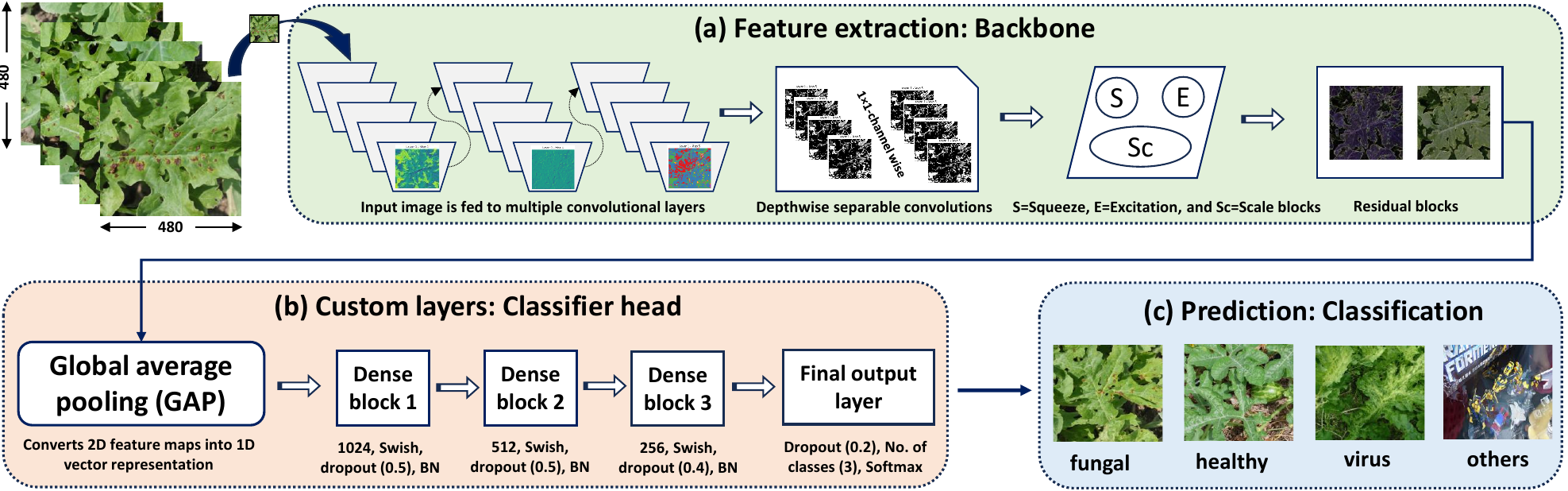}
    \vspace{-0.2cm}
    \caption{A schematic diagram representing EfficientNetV2-L model with various feature extraction blocks and layers. (a) feature extraction step where fine-grained features are extracted from images, such as edges, textures, and structure, (b) custom layer that was added to train the classifier head with the addition of three dense blocks and one final output layer for probability mapping for classes, and (c) prediction of test images into its four respective classes.}
    \label{fig4}
\end{figure}

In figure~\ref{fig4}a, the EfficientNetV2-L backbone which is trained on ImageNet was employed as a feature extractor. Within the feature extraction via backbone, the first few layers were kept frozen and only deeper layers were used to extract high-level image information. For instance, while low-level features, such as edges and textures were extracted through a CNN, the depth-wise separable convolutions were used to downsize the input image samples while preserving mid-level features. The subsequent layer, squeeze (S), and excitation (E) blocks were used to extract very specific fine-grained disease regions within the input images. In addition to this, a custom classifier head was then added on top of the backbone model, comprising three fully connected dense layers with Swish activations, batch normalization, and dropout regularization to enhance learning stability and prevent overfitting (Fig.~\ref{fig4}a).~The custom classifier was first trained independently while the base model remained frozen, ensuring that only the newly added layers learned data-specific disease features (Fig.~\ref{fig4}b).~Once the classifier reached a stable accuracy, fine-tuning was initiated on the deeper layers of EfficientNetV2-L architecture. This step allowed more room for improvisation on learning disease specific features while preventing unnecessary disruption of weights that may have caused overfitting or model instability.~All models were trained on images resized to 480$\times$480 resolution.~A progressive learning rate was used with a lower learning rate during fine-tuning to ensure better convergence. The model was trained using AdamW optimizer and the overall architecture was further stabilized using early stopping and learning rate reduction callbacks to achieve generalization power of the model for disease features.

\subsection{Metrics to evaluate model performance}

To assess model performance trained on all the categories of the dataset, four key metrics were chosen. These were: precision, recall, F1-score, and confusion matrix. While precision, recall, F1-score, and weighted F1-score demonstrated the model’s efficiency in classifying individual classes, a confusion matrix displayed the number of test images that were correctly classified by the model (Eqs.~\ref{equ1}-\ref{equ4}). In addition to this, to better understand feature separability under varying distributions of dataset with synthetic and real-field images (Table~\ref{Table1}), two tests, t-distributed Stochastic Neighbor Embedding (t-SNE)~\citep{rousseeuw1987silhouettes}, and Uniform Manifold Approximation and Projection (UMAP)~\citep{davies2009cluster} were also reported. t-SNE is a nonlinear dimensionality reduction technique that is designed to help visualize localized high-dimensional large-scale datasets in 2D space-displaying its potential clusters. Unlike t-SNE, UMAP on the other hand, highlights both local and global structure visualization in 2D space, thereby highlighting class separability. Although class-based cluster visualization is helpful, quantitative measures for these clusters, Silhouette and Davies-Bouldin Index, were also reported. These are based on equation~\ref{equ5} \& \ref{equ6}, respectively,

\begin{equation}
    \text{Precision} = \frac{TP}{TP + FP} 
    \label{equ1}
\end{equation}

\begin{equation}
    \text{Recall} = \frac{TP}{TP + FN}
\label{equ2}
\end{equation}

\begin{equation}
    \text{F1-score} = \frac{2 \times Precision \times Recall}{Precision + Recall}
\label{equ3}
\end{equation}

\begin{equation}
    \text{Weighted F1} = \sum_{i=1}^{N} w_i \times \text{F1}_i
\label{equ4}
\end{equation}

\begin{equation}
    S(i) = \frac{b(i) - a(i)}{\max(a(i), b(i))}
    \label{equ5}
\end{equation}

\begin{equation}
    \text{DBI} = \frac{1}{n} \sum_{i=1}^{n} \max_{j \ne i} \left( \frac{\sigma_i + \sigma_j}{d_{ij}} \right)
    \label{equ6}
\end{equation}

\section{Experimental results}\label{sec3}

\subsection{Evaluating model performance metrics and clustering validation}

\begin{figure}[t]
    \centering
    \includegraphics[scale=0.45]{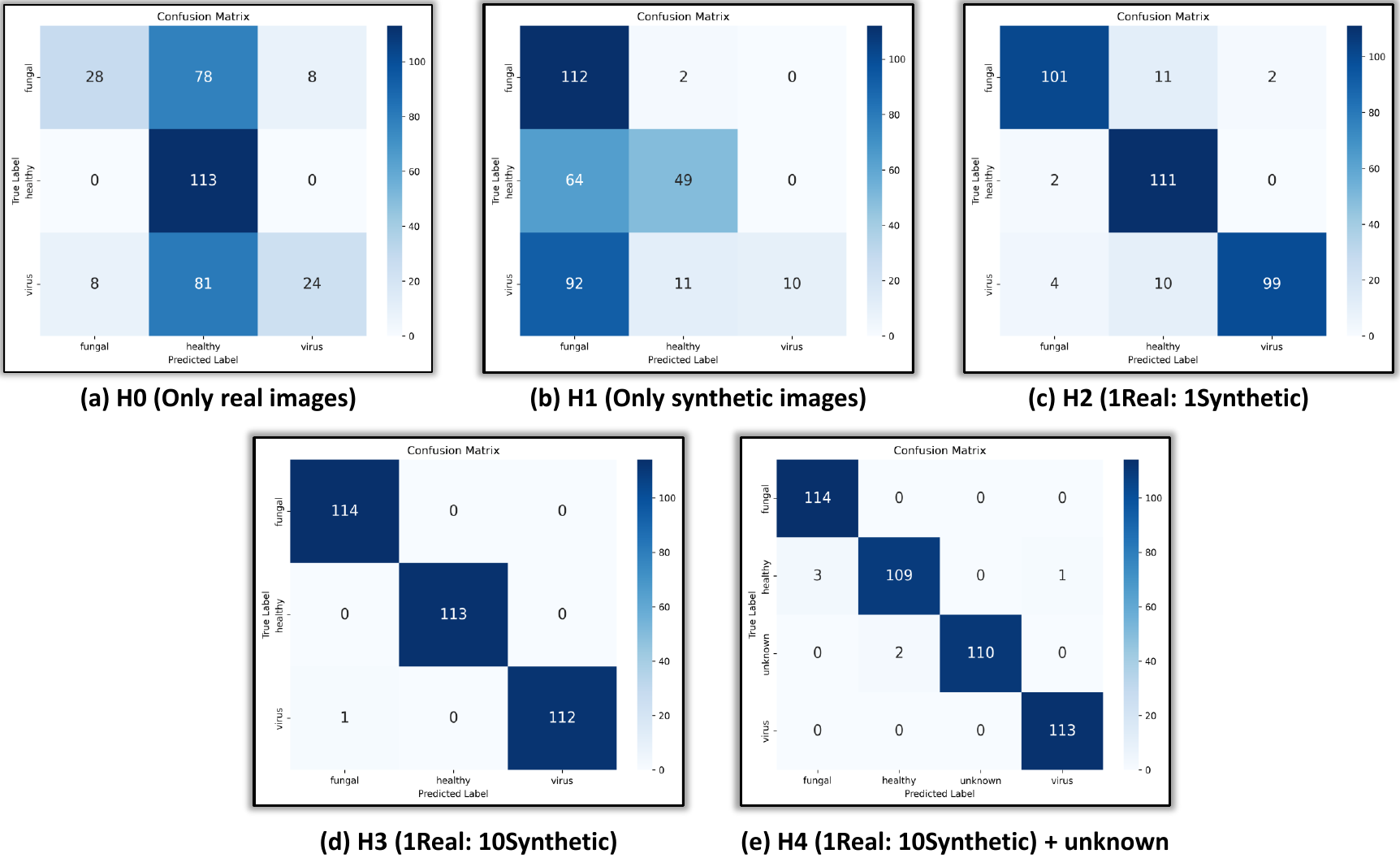}
    \vspace{-0.2cm}
    \caption{Confusion matrix demonstrating the number of classes with true labels vs. predicted ones for the following treatments: (a) H1 with only real images, (b) H2 with only synthetic images, (c) H2 with real to synthetic ratio of 1:1, (d) H3 with real to synthetic ratio of 1:10, and (e) H4 which is H3 with an addition of a random (unknown) class.}
    \label{fig5}
\end{figure}

The model demonstrated high precision, recall, and F1 scores for all classes in treatments H2-H4 (Table~\ref{Table2}). Test images of healthy samples are classified with nearly 100\% accuracy in almost all the models, except when only synthetic images were used (Fig.~\ref{fig5}b). Under the synthetic-only condition, healthy and viral symptoms were frequently misclassified as fungal symptoms, indicating limited model generalization. Notably, fungal classification improved (with 112 correctly classified) compared to the real images only (H0) condition, but 64 healthy and 92 virus images were misclassified as fungal, suggesting a bias toward fungal predictions. Another possible reason for the model's poor generalization could be attributed to irregular brown lesions (caused by wind damage or environmental stress) which were classified as fungal disease. Treatment H2 (Fig.~\ref{fig5}c), that incorporated 1:1 ratio for real and synthetic images, had strong performance across all classes with an improvement in precision and reduced misclassifications compared to only real or synthetic treatments. The confusion between virus and fungal classes dropped significantly, with virus class achieving over 99 correct predictions and only 4 and 10 misclassifications as fungal and healthy, respectively.

\begin{table}[h]
\caption{Multiple performance metrics reported by training models on four distributions of test dataset.} \label{Table2}
\begin{tabular*}{\textwidth}{@{\extracolsep\fill}lcccc}
\toprule%
Classes & Precision & Recall & F1-score & Weighted average F1-score \\
\midrule
\multicolumn{5}{c}{\textit{H0 (only real images)}} \\
\midrule
fungal & 0.78 & 0.25 & 0.37 \\
healthy & 0.42 & 1.00 &0.59 & 0.65 \\
virus & 0.75 & 0.21 & 0.33 \\
\midrule
\multicolumn{5}{c}{\textit{H1 (only synthetic images)}} \\
\midrule
fungal & 0.42 & 0.98 & 0.59 & \\
healthy & 0.79 & 0.43 & 0.56 & 0.74 \\
virus & 1.00 & 0.09 & 0.16 & \\ 
\midrule
\multicolumn{5}{c}{\textit{H2 (1real + 1synthetic)}} \\
\midrule
fungal & 0.94 & 0.89 & 0.91 & \\
healthy & 0.84 & 0.98 & 0.91 & 0.92 \\
virus & 0.98 & 0.88 & 0.93 & \\
\midrule
\multicolumn{5}{c}{\textit{H3 (1real + 10synthetic)}} \\
\midrule
fungal & 0.99 &1.00 & 1.00 & \\
healthy & 1.00 & 1.00 & 1.00 & 1.00 \\
virus & 1.00 & 0.99 & 1.00 & \\
\midrule
\multicolumn{5}{c}{\textit{H4 (1real + 10synthetic + unknown)}} \\
\midrule
fungal & 0.97 & 0.99 & 0.98 & \\
healthy & 1.00 & 0.99 & 1.00 & 0.99 \\
virus & 0.99 & 1.00 & 1.00 & \\
unknown & 1.00 & 0.97 & 0.99 & \\
\botrule
\end{tabular*}
\end{table}

When a small number of real images were combined with large synthetic images (H3), all classes were classified with near-perfect accuracy, including 114/114 fungal, 113/113 healthy, and 112/113 virus predictions, showing only a single instance of virus misclassified as fungal (Fig.~\ref{fig5}d). This demonstrates that when the model was trained with a mix of real and synthetic data that it generalized well across all classes. Finally, in the four-class classification treatment (H4), the addition of an unknown class, preserved the overall model performance (Fig.~\ref{fig5}e). The model successfully extends to the additional class (unknown) with high accuracy (110/112 for unknown) and almost prefect predictions for the remaining class. Overall, the results underscore the robustness of the trained EfficientNetV2-L architecture across diverse treatments. Additionally, even a small fraction of real images significantly improved the model performance specifically in terms of generalizability, making it well-suited for deployment in real-world disease classification scenario. 

In treatment H0 with only real images, the model exhibits poor generalization across classes, particularly for fungal and virus categories. While the healthy class shows perfect recall (1.00), its low precision (0.42) indicates a high number of false positives which is likely due to the model over-predicting this class (Table~\ref{Table2}). The fungal and virus classes suffer from low recall scores (0.25 and 0.21, respectively), suggesting that the model struggles to correctly identify these classes. The overall weighted F1-score was 0.65, highlighting that the model struggles to generalize when trained on a limited quantity of real images alone due to inadequate image diversity and insufficient representation of visual patterns. Fungal recall remains high at 0.98, but the precision drops to 0.42 suggesting a high rate of false positives. This could be possibly due to synthetic data over-representing features associated with fungal symptoms. Virus classification remains extremely poor with recall of 0.09 and F1-score of 0.16, likely because synthetically generated virus symptoms does not adequately capture real-field variations (Table~\ref{Table2}). The healthy class shows moderate performance with a precision of 0.79 and recall of 0.43. The overall weighted F1-score improved to 0.74 compared to H0 treatment, showing that while generating synthetic images increase training volume, it cannot replace real images or introduce better generalization in real-field conditions. 

The model trained with a 1:1 ratio of real to synthetic images performed more accurately in terms of performance metrics. All three classes, fungal, healthy, and virus, achieve high precision and recall values (all above 0.84), with F1-score ranging from 0.91-0.93. This balance between real and synthetic images enables the model to learn more robust, generalizable features~\citep{bafghi2025mixdiff}. In comparison to using only real or synthetic images, he weighted F1-score increased to 0.92, indicating effectiveness of combing both real and synthetic images. Using a 1:10 ratio of real to synthetic images resulted in perfect classification, with precision, recall, and F1-scores all equal to 1.00 across all three classes. The weighted F1-score of 1.00 illustrates accuracy. Finally, the model trained on the H4 treatment with the same 1:10 real-to-synthetic ratio used in H3, but with the inclusion of an additional unknown class, handled the added complexity exceptionally well, thereby achieving near-perfect precision and recall across all the classes. F1-scores remain above 0.97, including a strong 0.99 for the newly introduced unknown class. The virus class, which previously showed inconsistencies in H0-H2, achieves perfect classification (Table~\ref{Table2}). The weighted F1-score is 0.99, suggesting that the model not only adapts to an expanded class structure but does so with minimal loss in accuracy or generalization.

\begin{figure}[h]
    \includegraphics[scale=0.46]{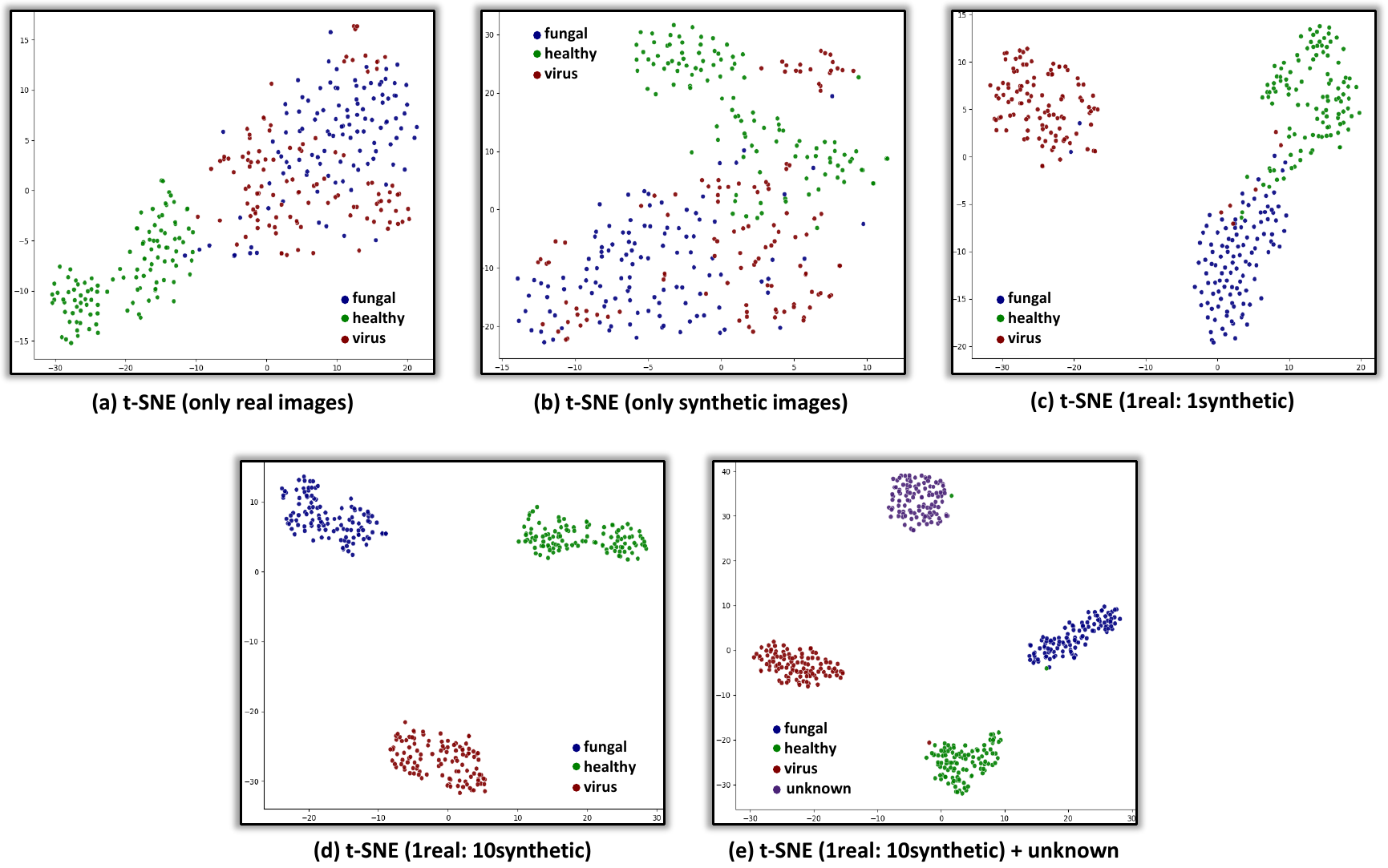}
    \vspace{-0.2cm}
    \caption{The 2D t-distributed Stochastic Neighbor Embedding (t-SNE) plot shows the model’s test results, with colors representing classes:~(\textcolor{blue}{fungal}, \textcolor{green}{healthy}, \textcolor{red}{virus}, and \textcolor{violet}{unknown}).}
    \label{fig6}
\end{figure}

The subsequent plots display feature embeddings for t-SNE based on high-dimensional representations extracted by EfficientNetV2-L models (Fig.~\ref{fig6} a-e). Each plot shows the 2D projections of high-dimensional features, with points color-coded according to the multi-class classification. The treatment using only real-images (H0) exhibited more compact clusters (especially healthy class) compared to the treatment using only synthetic images (Fig.~\ref{fig6}a). This is consistent with the lowest Silhouette and highest DBI scores, suggesting poor intra-class compactness and significant overlap between class boundaries (Table~\ref{Table3}). Using only synthetic data shows more scattered plot with lower Silhouette score (Fig.~\ref{fig6}b \& Table~\ref{Table3}). A 1:1 ratio of real to synthetic data improved clustering significantly with a Silhouette and DBI scores of 0.61 and 0.49 (Fig.~\ref{fig6}c \& Table~\ref{Table3}). This combination highlights the value of both image types. Real images capture diverse morphologies and textures with authentic symptom appearance, while synthetic images provide consistent representations, together improving intra-and-inter class separations. The use of a 1:10 real-to-synthetic (H3) data achieved the best clustering outcome (Fig.~\ref{fig6}d \& Table~\ref{Table3}). Additionally, including an unknown class (H4) did not negatively affect clustering (Silhouette and DBI scores of 0.81 and 0.27, respectively), demonstrating the model’s ability to learn boundaries even in the presence of non-relevant images (Fig.~\ref{fig6}). This is important for real-time disease classification where unknown objects, such as weeds, plastic mulch, brown lesions caused by strong winds or environmental factors often pose challenges in accurately classifying disease symptoms.

\begin{figure}[h]
    \includegraphics[scale=0.46]{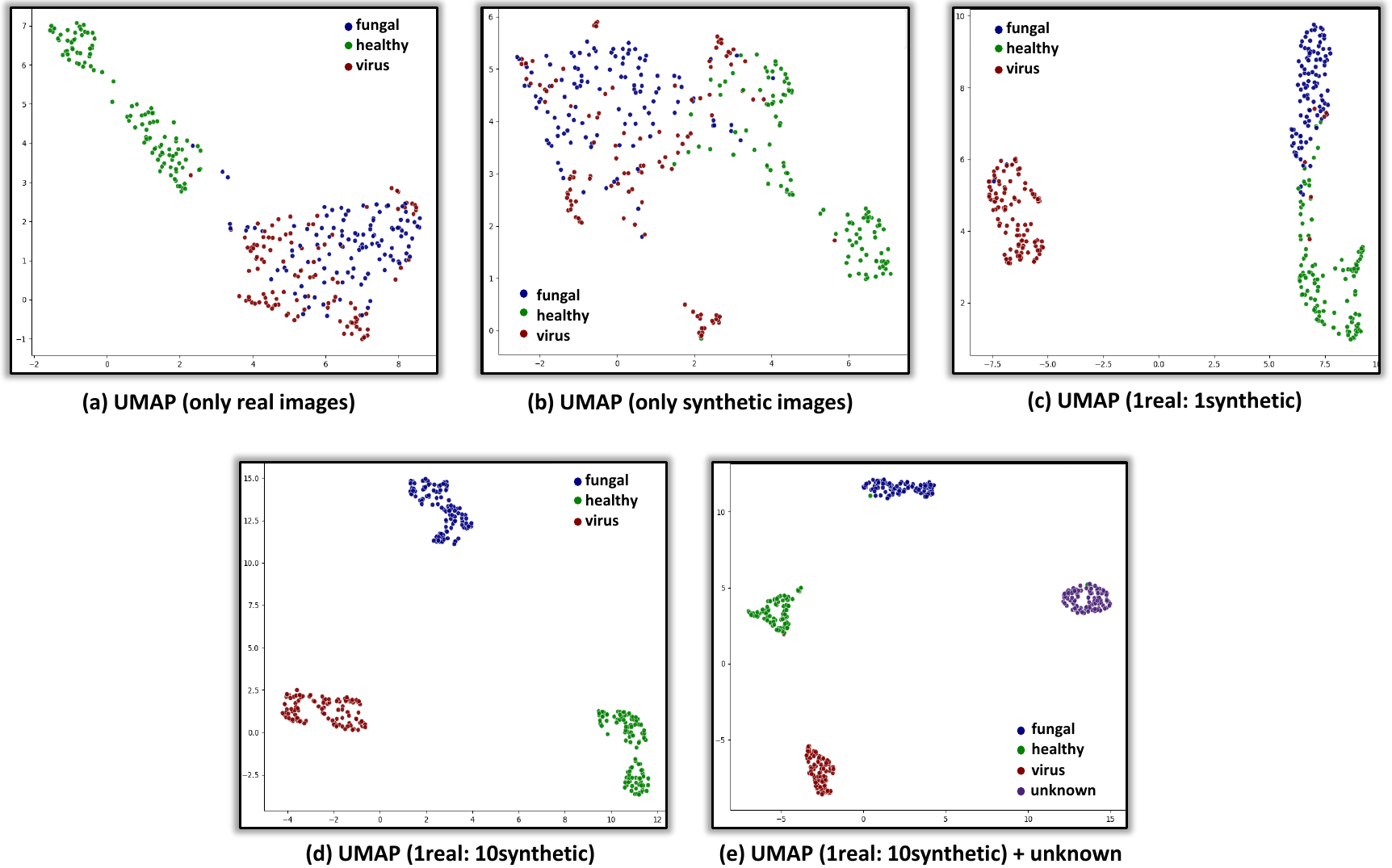}
    \vspace{-0.2cm}
    \caption{The 2D Uniform Manifold Approximation and Projection (UMAP) plot shows the model's test results, with colors representing classes:~(\textcolor{blue}{fungal}, \textcolor{green}{healthy}, \textcolor{red}{virus}, and \textcolor{violet}{unknown}).}
    \label{fig7}
\end{figure}

In figure~\ref{fig7}, UMAP projections are presented for the feature embeddings learned by EfficientNetV2-L models across all treatments. UMAP, unlike t-SNE, preserves both local and global feature structure, thereby providing better insights into how effectively the model separates and organizes class features in a lower-dimensional space. When only real images were used, the model exhibited poor clustering, as reflected by low Silhouette and DBI scores (Table~\ref{Table3}). Similarly, the use of only synthetic images produced comparable patterns (Fig.\ref{fig7}b). Clustering quality improved substantially when a 1:1 ratio of real-to-synthetic images was used, achieving Silhouette and DBI scores of 0.65 and 0.45, respectively. This balance reaffirms the t-SNE findings regarding the model's ability to generalize learned features to the test set. The best results were obtained in treatments where real images were added to a substantially larger synthetic set (Figs.\ref{fig7}d \& e, Table~\ref{Table3}).~Overall, both t-SNE and UMAP analyses reinforce the conclusion that mixed datasets, particularly in the ratios used for H3 and H4, enable the model to extract more meaningful features than when using either real or synthetic data alone. Despite the lower clustering metrics observed with only real images, the inclusion of real samples remains essential for accurate disease classification.

\begin{table}[h]
\caption{Clustering quantitative metrics (Silhouette and Davis-Bouldin coefficients) for t-SNE and UMAP across five treatment distribution.} \label{Table3}
\begin{tabular*}{1\textwidth}{@{\extracolsep\fill}lcccc}
\toprule%
Treatments & \multicolumn{2}{c}{t-SNE} & \multicolumn{2}{c}{UMAP} \\\cmidrule{2-3}\cmidrule{4-5}%
& Silhouette score\textsuperscript{[1]} & DBI\textsuperscript{[2]} & Silhouette score\textsuperscript{[1]} & DBI\textsuperscript{[2]} \\
\midrule
H0  & 0.29 & 2.28 & 0.31 & 1.79 \\
H1 & 0.16 & 1.99 & 0.21 & 2.04 \\
H2 & 0.61 & 0.49 & 0.65 & 0.45 \\
H3 & 0.82 & 0.26 & 0.85 & 0.21 \\
H4 & 0.81 & 0.27 & 0.86 & 0.21 \\
\botrule
\end{tabular*}
\footnotetext[1]{Higher is better. \footnotetext[2]{Lower is better.}} 
\end{table}

\section{Discussion} \label{Sec4}

\subsection{Synthetic images as a substitute for real images}

With the current need for developing mapping and targeted spraying technology that involves generating and training large-scale datasets, synthetic images generated using GenAI approaches could be an alternative. However, it does not necessarily validate the replacement of real images collected via various sensing systems (RGB digital camera in this study). Real images bring a plethora of variations within the image, such as different soil backgrounds, dynamic lighting conditions, unpredictable weather situations, occluded plant structures, and unpredictable variation (dust, motion blur with wind, or dew). In addition, real images may also capture sensor related distortions, such as lens artifacts or shutter effects, which GenAI model might fail to capture when generating synthetic images. Synthetic images improve visual clarity and reduce intra-class separability in latent space; however, real images contribute to essential noise and necessary phenotypic realism necessary for model generalization in real-field conditions~\citep{singh2024synthetic,liu2022detecting}. Therefore, while synthetic images add value to large-scale dataset, they cannot replace or be used as a substitute for real images. In this case, a hybrid approach, merging real images with synthetic ones will enhance model performance in real-world conditions. 

\begin{figure}
    \includegraphics[scale=0.68]{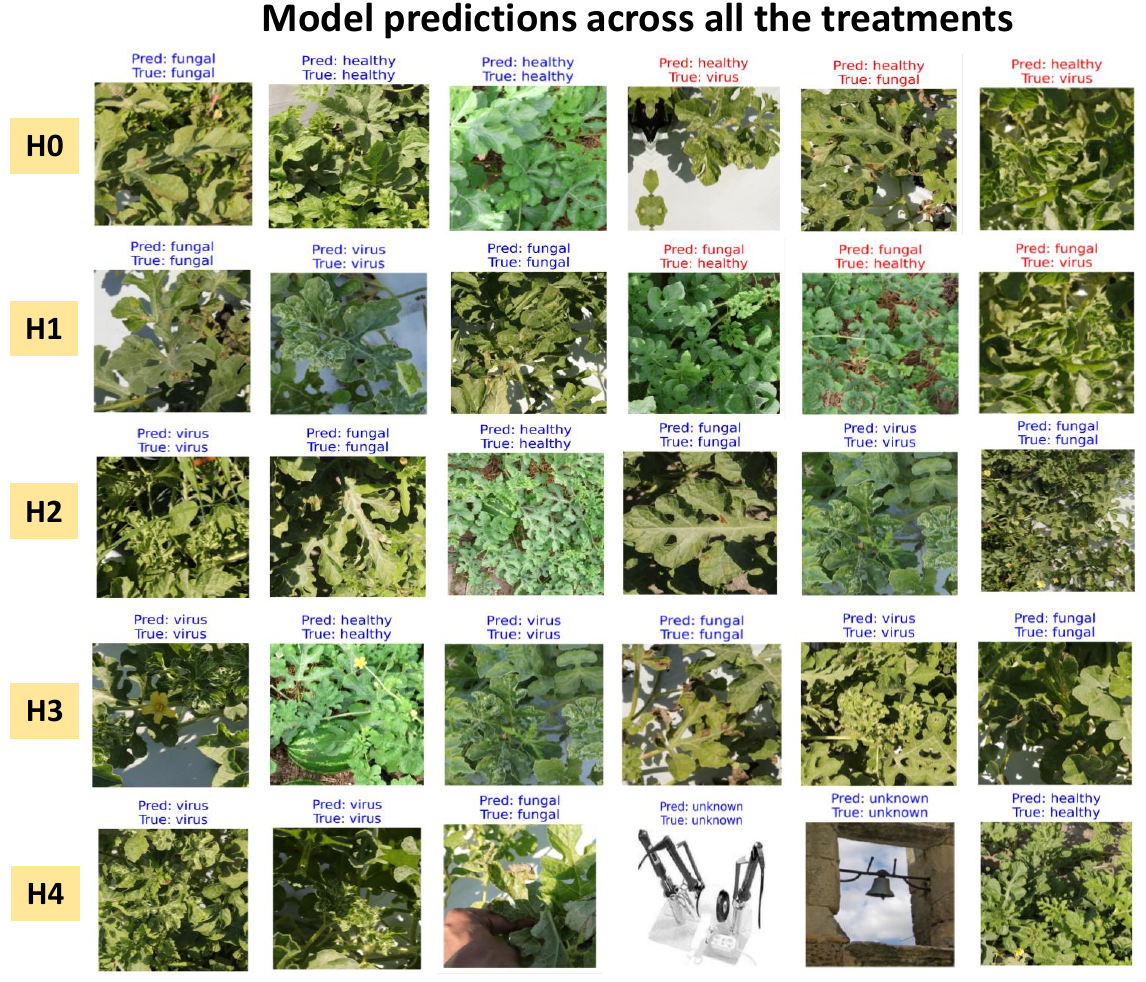}
    \vspace{-0.2cm}
    \caption{Classified test images with ground truth. “Pred” corresponds to prediction that the model made on the test set and “True” means ground truth labeled dataset. Following colors represent the predictions: \textcolor{blue}{blue} is correct and \textcolor{red}{red} is incorrect.}
    \label{fig8}
\end{figure}

Figure~\ref{fig8} presents classified test images from treatments H0-H4 using the trained EfficientNetV2-L model. As discussed in Section~\ref{sec2.1}, only real images were used for testing and evaluating the model’s generalizability. The model's prediction across different treatments reflects both strengths and challenges in real-world prediction scenarios. When only real images were used (H0), the model did not generalize well in the presence of shadow and brown lesions. For instance, in the last test image, virus symptoms were presented with brown lesions caused due to wind damage. Due to overfitting problem, the model did not learn virus features adequately and predicted it as healthy. Similar prediction could be seen when only synthetic images were used (Fig.~\ref{fig8}). This also highlights the difficulty in distinguishing between early or mild symptoms from visually similar healthy leaf patterns. For real-to-synthetic images (1:1), although the model correctly classified most images, environmental artifacts, such as shadow or mulch, introduced some inconsistencies. Notably, minute features of these classes were correctly learned by the model and correct predictions were made. H3 (real-to-synthetic, 1:10) shows strong performance with accurate predictions across all three classes. The model identified most of the test classes correctly even in the presence of mulch or overlapping leaves. This again suggests robust learning in this treatment, possibly due to high-quality real image representation during training~\citep{singh2024synthetic}. With the inclusion of an irrelevant class, the model rejects out-of-distribution images (unknown class) and predicts it accurately. Overall, while synthetic data played a crucial role during training, the test predictions underscore the importance of diverse real-field images for achieving robustness, particularly in handling noise, occlusion, and unknown objects in real-world scenario. 

\subsection{Model generalization when adding real images and an unknown class}

The addition of real images within the synthetic images imparted naturality that synthetic images failed to capture. This includes crop occlusion, motion blur, and spatial differences, such as camera angles and height. These are various ``edge-cases'' that helped the model to achieve better generalization on the test set. By integrating as little as 638 real images (in a ratio of 1:10 real-to-synthetic) captured in real-field conditions, the model was exposed to examples that helped it achieve better precision, recall, and F1-scores. The overall weighted F1-score increased from 0.65 to 1.00 (Table~\ref{Table2}). From a more practical point of view, adding an irrelevant (unknown) class also helped the model to achieve better results. For instance, the unknown class acted as a distractor mimicking broad domain conditions where the model must recognize not just the disease patterns, but also what it should recognize as something not belonging to any crop-centered category (common objects)~\citep{buda2018systematic}. Additionally, this was also evident in the F1-score and weighted F1-score of 1.00 and 0.99, respectively (Table~\ref{Table2}).  

\subsection{Distinct feature space}

The t-SNE metric tends to preserve local features better with a focus on intra-class separability. In this research, two necessary parameters to obtain t-SNE plots were used: perplexity (30) and ``n\_components'' (2). To obtain better clustering plots, perplexity is an important argument that tends to define the effective nearest neighbors. Its value typically ranges between 5-50; however, setting this value depends on the size of the dataset used. Since this research moderately used large number of input samples, perplexity score was set to 30 to understand distinct class separability. On the other hand, ``n\_components'' determines the visualization in 2D or 3D space, therefore, its value was set to 2 for 2D cluster visualization. Unlike t-SNE, UMAP tends to preserve both local as well as global structures of the dataset. Both t-SNE and UMAP preserved excellent intra-and-inter-class separability when  a combination of real and synthetic images were merged with a ratio of 1:1. Based on the representation of high dimensionality reduction clusters obtained using t-SNE and UMAP clustering tests, it is evident that combing synthetic with real images allow models to learn more resilient features which is a critical requirement for practical field deployment of these computer vision algorithms.

\section{Limitations and future research considerations} \label{Sec5}

Through this experimental study, several limitations were observed that could have impacted the overall metrics accuracy along with the model classification performance. Additionally, possibility for future research directions have also been added for interested researchers. Some of these limitations and future research directions are:
\begin{enumerate}
    \item \textbf{Uncertainty with the sample sizes of training and testing dataset:} Although a substantial dataset was trained for model training and testing on unseen samples, it remains unclear whether increasing the number of images, either in the synthetic or real training set, would improve model performance. The current experimental design only provides a fixed subset of real images (750 in H0), leaving the effect of sub-optimal real images on overall model performance uncertain. Future studies must consider scaling up both synthetic and real datasets or incrementing the real images in smaller steps, such as in the ratio of 1:1, 1:2, 1:3, and so on up to 1:10. This should be accomplished in progressive manner rather than skipping directly from 1:1 to 1:10, to more precisely evaluate their combined impact on model robustness and real-world generalization.
    \item \textbf{Lack of external validation and testing:} The developed models were tested on images from a single experimental site. Broader utility requires external validation using independently collected samples from diverse regions, seasons, management practices, and imaging devices, such as those from commercial farms with multiple disease symptoms. 
    \item \textbf{Unrealistic coloration of synthetic images:} The synthetic images used in this study had inconsistency with coloration of the diseased leaves. For instance, in figure~\ref{fig2}, the real images (right box) were captured in sunny conditions and where the leaf colors were mostly light green. However, most of the synthetic images were dark green in color. This may have affected the model accuracy which was not assessed in this study. This may have led to significant domain shift from synthetic samples in training set to real images in the test set leading to less generalization under natural color variations.
    \item \textbf{Multiple disease combinations for synthetic image generation:} In this research study, the training examples to train SD 3.5M model only consisted of single classes of disease to generate synthetic samples. For instance, in figure~\ref{Figure3}, only anthracnose or downy mildew disease classes were used to generate its respective synthetic samples. However, in real-field conditions that may not be the case and multiple disease symptoms could appear on leaf surfaces. In this case, it becomes challenging for SD 3.5M model to capture specific information about multiple disease symptoms and generate authentic disease examples. This is one of the existing challenges of GenAI-based text-to-image multi-modal models where image-to-image translation might be a better fit. 
\end{enumerate}

\section{Conclusion} \label{Sec6}

 This research study presents a thorough evaluation of testing the efficacy of GenAI-based synthetic images in enhancing watermelon \textit{(Citrullus lanatus)} disease classification when integrated with real images. To carry out this analysis, five treatments were evaluated. These were: H0 (only real images), H1 (only synthetic images), H2 (1:1 real-to-synthetic), H3 (1:10 real-to-synthetic images), and H4 (H3 + unknown class). The overall treatments were trained using an EfficientNetV2-L model with multiple fine-tuning and transfer learning approaches. 

Adding a smaller number of real images with synthetic images appeared to increase overall model performance and generalizability. This was evident where the ratios of real-to-synthetic images was 1:1 and 1:10, respectively. Not only the metrics performance increased, but it also boosted the overall performance of the trained EfficientNetV2-L model in terms of generalizability and feature class separability. Based on the cluster analysis and quantitative measure, it was evident that synthetic images cannot substitute for real images. In other words, although synthetic images could be generated in large numbers using various GenAI open-access models, real images acquired using various sensing system still carry importance in terms of leaf morphology, natural variability in symptom expression, and sensor-specific noise~\citep{geng2024unmet,singh2024synthetic}. Therefore, CNN models trained specifically on synthetic datasets may suffer from these limitations and fail to adequately classify diseases in real-field settings due to domain shift. 

\section*{CRediT authorship statement}
\textbf{Nitin Rai:}~Data curation, Formal analysis, Investigation, Methodology, Writing-original draft.~\textbf{Nathan S. Boyd:}~Resources, Project administration, Validation, Writing-review \& editing.~\textbf{Gary E. Vallad:}~Resources, Project administration, Writing-review \& editing.~\textbf{Arnold W. Schumann:}~Conceptualization, Methodology, Supervision, Writing-review \& editing.

\section*{Declaration of competing interest}
The authors declare no financial or conflicting interests in this research. 

\section*{Acknowledgment}
This research was supported by the United States Department of Agriculture (USDA)-Small Business Innovation Research \& Technology Transfer Programs (SBIR/STTR) grant \# 2024-51402-42007. Thanks to Emily Witt and Michael Sweat for their assistance with field experiments.

\bibliography{sn-bibliography}

\end{document}